\documentclass[10pt,twocolumn,letterpaper]{article}

\usepackage{iccv}
\usepackage{times}
\usepackage{epsfig}
\usepackage{graphicx}
\usepackage{amsmath}
\usepackage{amssymb}

\usepackage{amsmath,amssymb,amsfonts}
\usepackage{algorithmic}
\usepackage{graphicx}
\usepackage{float}
\usepackage{textcomp}
\usepackage{multirow}
\usepackage{xcolor}
\usepackage{ulem}
\usepackage{balance}
\usepackage{subfigure}
\usepackage{lipsum}

\newcommand\blfootnote[1]{%
  \begingroup
  \renewcommand\thefootnote{}\footnote{#1}%
  \addtocounter{footnote}{-1}%
  \endgroup
}
\usepackage[breaklinks=true,bookmarks=false]{hyperref}

\iccvfinalcopy 

\def\iccvPaperID{****} 

\ificcvfinal\fi

\begin{document}

\title{Can Self-Supervised Representation Learning Methods Withstand Distribution Shifts and Corruptions?}
\author{Prakash Chandra Chhipa\textsuperscript{1,*}, Johan Rodahl Holmgren\textsuperscript{1}, Kanjar De\textsuperscript{1,2,**}, Rajkumar Saini\textsuperscript{1} and Marcus Liwicki\textsuperscript{1}\\
\textsuperscript{1}Machine Learning Group, EISLAB, Lule\aa~Tekniska Universitet, Lule\r{a}, Sweden\\
\textsuperscript{2}Video Coding Systems, Fraunhofer Heinrich-Hertz-Institut, Berlin, Germany\\
{\tt\small [first].[middle].[last]@ltu.se}\\
\textsuperscript{*}{\tt\small corresponding author - prakash.chandra.chhipa@ltu.se}
}

\maketitle
\ificcvfinal\thispagestyle{empty}\fi

\begin{abstract} 
Self-supervised representation learning (SSL) in computer vision aims to leverage the inherent structure and relationships within data to learn meaningful representations without explicit human annotation, enabling a holistic understanding of visual scenes. 
Robustness in vision machine learning ensures reliable and consistent performance, enhancing generalization, adaptability, and resistance to noise, variations, and adversarial attacks. Self-supervised representation learning paradigms, namely contrastive learning, knowledge distillation, mutual information maximization, and clustering, have been considered to have shown advances in invariant learning representations.
This work investigates the robustness of learned representations of SSL approaches focusing on distribution shifts and image corruptions in computer vision. Detailed experiments have been conducted to study the robustness of SSL methods on distribution shifts and image corruptions. The empirical analysis demonstrates a clear relationship between the performance of learned representations within SSL paradigms and the severity of distribution shifts and corruptions. Notably, higher levels of shifts and corruptions are found to significantly diminish the robustness of the learned representations. These findings highlight the critical impact of distribution shifts and image corruptions on the performance and resilience of SSL methods, emphasizing the need for effective strategies to mitigate their adverse effects. The study strongly advocates for future research in the field of self-supervised representation learning to prioritize the key aspects of safety and robustness in order to ensure practical applicability.
The source code and results are available on GitHub.
\footnote{{\href{https://github.com/prakashchhipa/Robsutness-Evaluation-of-Self-supervised-Methods-Distribution-Shifts-and-Corruptions} {https://github.com/prakashchhipa/Robsutness-Evaluation-of-Self-supervised-Methods-Distribution-Shifts-and-Corruptions}}}.
\blfootnote{\textsuperscript{**}\textit{Work performed at Machine Learning Group, EISLAB, Lule\aa~Tekniska Universitet, Lule\r{a}, Sweden}}
\end{abstract}

\section{Introduction}
\label{sec:Introduction}

Safety and robustness are crucial in computer vision as they ensure the accurate and reliable perception of the visual world, enabling applications such as autonomous driving~\cite{li2022self}, and surveillance systems to make informed and trustworthy decisions, reduce environmental noise~\cite{he2021reducing}, ultimately enhancing overall human safety and well-being.
In recent years self-supervised representation learning (SSL) methods~\cite{ericsson2022self} have garnered interest in computer vision applications. Its current state-of-the-art is prominent even against supervised examples where invariant representation learning has been the core, as stated in~\cite{ericsson2022selfbmvc}. SSL is a well-explored representation learning approach, with many studies on its performance in large datasets such as ImageNet-2012 and also on multi-modality~\cite{zong2023self}.
In addition, SSL has also been well-explored with other learning approaches, including active learning~\cite{caramalau2022mobyv2}, graphs~\cite{shirian2022self}, life-long learning~\cite{thota2022lleda}, and many more.
Recent advances in self-supervised representation learning can be broadly categorized into multiple paradigms, namely contrastive learning \cite{simclr,moco}, Knowledge Distillation \cite{byol,dino,simsam}, Mutual Information Maximization \cite{barlowtwin,bardes2022vicreg}, and Clustering \cite{swav}.
Despite these advancements, the robustness and safety aspects of SSL paradigms have not been extensively explored, which hinders their applicability in real-world use cases.
This study is one of the early attempts highlighting the above-stated research gap on a large-scale dataset \cite{imagenetc} focusing on the distribution shifts and image corruptions.
Representation learning from self-supervised representation learning paradigms for computer vision can be categorized majorly as (i) Joint Embedding Architecture \& Method (JEAM) (\cite{chen2020simple}, \cite{grill2020bootstrap}, \cite{caron2020unsupervised}, \cite{zbontar2021barlow}), (ii) Prediction Methods (\cite{veeling2018rotation}, \cite{noroozi2016unsupervised}, \cite{doersch2015unsupervised}), and loosely (iii) Reconstruction Methods~(\cite{kingma2013auto}, \cite{goodfellow2014generative}).
Specifically, JEAM can be divided further with each subdivision providing many interesting works; (i)~Contrastive Methods (PIRL~\cite{misra2020self}, SimCLR~\cite{chen2020simple}, SimCLRv2~\cite{chen2020big}, MoCo~\cite{he2020momentum}), (ii)~Distillation (BYOL~\cite{grill2020bootstrap}, SimSiam~\cite{chen2021exploring}), (iii)~Quantization (SwAV~\cite{caron2020unsupervised}, DeepCluster~\cite{caron2018deep}), and (iv)~Information Maximization (Barlow Twins~\cite{zbontar2021barlow}, VICReg~\cite{bardes2021vicreg}).
Robustness is critical in real-life computer vision applications as there will be a shift in distribution for the deployed models with time. Understanding the behavior of existing models to the distribution shift is a crucial consideration in developing newer, more robust models.

Ericsson et al. \cite{ericsson2022selfbmvc} explore the impact of different augmentation strategies on the transferability of self-supervised representation learning models to downstream tasks. The authors show that CNNs trained contrastively do learn invariances corresponding to the augmentations used, and specializing CNNs to particular appearance/spatial augmentations can lead to greater corresponding invariances. Furthermore, learning invariances to synthetic transforms does provide a degree of invariance to corresponding real-world transforms. This work establishes the correspondence between synthetic transforms and learning invariances for knowledge transfer limited to ~\cite{imagenet} without focusing on robustness and distribution shift.

Another significant work by Jiang et al.~\cite{jiang2020robust} focuses on improving the robustness of self-supervised pre-training by learning representations that are consistent under both data augmentations and adversarial perturbations. It leverages contrastive learning to enhance adversarial robustness via self-supervised pre-training. They discuss several options to inject adversarial perturbations to reduce adversarial fragility. Through experiments in both supervised fine-tuning and semi-supervised learning settings, they demonstrate that the proposed adversarial contrastive learning can lead to models that are both label-efficient and robust. The paper does not specifically focus on corruption, but rather on improving the model's ability to handle adversarial attacks. This work shows notable improvement in robustness performance but remains limited to a small-scale CIFAR dataset, subject to limited generalizability.

Research is needed to learn invariant SSL representations capable of handling distribution shifts and corruptions; this study provides a ground in this direction by sharing insights into the robustness performance of a large-scale dataset. The identified research gap(s), raises several research questions addressed in later sections. For the detailed investigation, we considered the most popular SSL paradigms, namely contrastive learning, knowledge distillation, mutual information maximization, and clustering. Next, we exhaustively evaluated the corruptions and their severity levels present in ImageNet-C dataset~\cite{imagenetc} to understand the resilience of each method. Further, compare the robustness performance across multiple metrics, including qualitative analysis. To the best of our knowledge, this is one of the early works in this direction.

\textbf{Q1}: \textit{How do self-supervised representation learning (SSL) paradigms (contrastive learning, knowledge distillation, mutual information maximization, clustering) perform in terms of robustness when exposed to distribution shifts and image corruptions?} 
A1: Distribution shifts and image corruptions have an effect on the robustness performance of the well-known SSL paradigms. The empirical analysis in this study shows that the error rates (averaged over all distribution shifts and image corruptions) increase with an increase in the severity levels of the distribution shifts and image corruptions. (Figure \ref{fig:avg}, and Section~\ref{Q1}.Q1). 



\textbf{Q2}: \textit{To what extent can self-supervised representation learning methods maintain their robustness in the presence of distribution shifts, and what are the factors that limit their ability to do so?} 
A2: Extensive experiments reveal that SSL methods sustain robustness performance 
 when subjected to lower levels of corruptions, and subsequently, the performance reduces when subjected to higher levels of corruptions. Higher corruptions may lead to massive distribution shifts, which may affect the robustness performance of learned representations.  (Figure \ref{fig:3-elst-sat-sno}, Table~\ref{tab:overview_large}, and Section~\ref{Q2}.Q2). 

\textbf{Q3}: \textit{What is the relationship between the robustness of different SSL paradigms and common categories of corruptions?}   A3: Generally, robustness performance decreases for increased severity of corruptions; specifically, the weather group's robustness performance is poorer than that of other groups. (Figure \ref{fig:group}, and Section~\ref{Q3}.Q3).

\textbf{Q4}: \textit{Do self-supervised representation learning methods deviate from the observed trend of error increase for certain corruptions, and what factors contribute to their robustness in the face of these corruptions?} 
A4: Yes; a few corruptions, namely, \textit{snow, elastic transform}, and \textit{saturate}, deviate from the observed trend supported by visual quality analysis. (Table~\ref{tab:ssim}, and Section~\ref{Q4}.Q4).

\textbf{Q5}: \textit{To what extent does the presence of corruptions shift the focus of classifiers from overall representation to specific features?} 
A5: GradCam~\cite{gradcam} analysis reveals that there is a significant shift in the attention maps when the image is subjected to higher levels of corruption. (Figure~\ref{fig:grad_dog_grid}\&\ref{fig:grad_rep_grid}, and Section~\ref{Q5}.Q5).

\textbf{Q6}: \textit{Do different backbones, such as Convolutional Neural Networks (CNNs) and Transformers, influence the behavior and robustness?} 
A6. Yes; the self-attention mechanism in transformer, in contrast to CNNs, does not embed any visual inductive bias of spatial locality \cite{jelassi2022vision}. (Figure~\ref{fig:dino_compare_ic}, and Section~\ref{Q6}.Q6).

\section{Methodology }

Comparative performance evaluation against robustness is carried out in two steps. In the first step, self-supervised representation learning method(s) are chosen from each potential self-supervised representation learning approach (based on JEAM), including contrastive learning, knowledge distillation, mutual information maximization, and clustering. In the second step, evaluation measures are chosen, indicating quantitative and qualitative comparisons on distribution shift and corrupted data samples from ImageNet-C.

\textit{Reason for measuring robustness of learnt representations with corruptions and severity} - This study focuses specifically on robustness of representations where domain shifts is simulated in controlled manner through corruption and their varying severity level.
Corruptions and perturbations in ImageNet-C~\cite{imagenetc} are meticulously curated and carefully designed to closely simulate natural phenomena in vision, related to geometric distortions, visual noises, and other explicit factors. Five severity levels further resembles the increased difficulty level, aiding to study robustness at scale. Corruptions across multiple severity levels, thereby altering the original data distribution in a controlled manner~\cite{imagenetc}. Each corruption severity level shifts the distribution progressively. The corruptions cause variations in texture, color, and spatial coherence, effectively expanding the data manifold towards shift. 
\subsection{Self-supervised Representation Learning Methods}
Methods from different self-supervised representation learning approaches are considered for analysis on the ImageNet-C dataset. The self-supervised representation learning techniques considered for this work are categorized into four main categories based on their methodology.

\textbf{Contrastive Learning}: It is a self-supervised representation learning approach in computer vision and other machine learning domains. The principle behind contrastive learning is to learn valuable representations by encouraging similarity between semantically similar data points while maintaining dissimilarity between unrelated or contrasting data points. In computer vision, this approach helps in learning features and representations from images without relying on labeled data. Instead, it exploits the inherent structure in the data to learn meaningful representations that can be used for various downstream tasks. Specifically, SimCLR method \cite{simclr} minimizes the temperature-scaled loss function. This contrastive loss penalizes the network when positive pair similarity is low and negative pair similarity is high.

\textbf{Knowledge Distillation}: Distillation-based self-supervision is where student and teacher style encoders are structured and share the learning weights with specific arrangements such as exponential moving averages. Typically, similarity learning is performed by inducing architectural dissimilarity, such as adding a prediction MLP network on only one of the branches. In this work, SimSiam \cite{simsam}, a self-distillation method, and BYOL \cite{byol} \& DINO (with ResNet encoder) \cite{dino} dual encoder style knowledge distillation methods are employed.

\textbf{Mutual Information Maximization}: This principle is used in self-supervised representation learning to learn valuable and meaningful representations from data without explicit labels. The principle is to maximize the mutual information between different views or transformations of the input data, assuming that the learned representations should be invariant or robust to these transformations. Barlow Twins \cite{barlowtwin}, and VICReg \cite{bardes2022vicreg} are two self-supervised representation learning methods employed for the work to follow the principle of mutual information maximization to learn visual representations by applying redundancy reduction.

\textbf{Clustering}: SwAV \cite{swav} combines contrastive learning and clustering-based approaches to learning meaningful and invariant features from images. The main idea behind SwAV is to use a clustering mechanism to enforce consistency between different views of the same image while promoting diversity in the learned representations.

\textbf{Robustness Evaluation Criteria}: The error rate metrics, namely corruption error ($CE$), mean corruption error ($mCE$), clean error, average error, and average relative error, were introduced as a standardized measure to benchmark the robustness of machine learning models on Imagenet-C. The two-step evaluation is described by Hendrycks et al.~\cite{imagenetc}. The same procedure has been followed in this study.  

\subsection{Dataset and Experimental Setup}

\textbf{ImageNet-C} dataset~\cite{imagenetc} contains 19 types of corruptions with five severity levels, each algorithmically generated. The main objective is to analyze the performance of different self-supervised representation learning methods across these corruptions and severity levels. By conducting detailed experiments, this research aims to gain insights into how self-supervised representations handle various types of corruptions. In this paper, we have performed detailed experiments considering all the corruptions and severity levels to gain a deeper understanding of how different self-supervised representations work on different types of corruptions and present our findings in Section~\ref{sec:RQ}.

\begin{table}[h!]
 \centering
 \caption{Configuration used (refer \cite{mmselfsup2021,dino} for implementation details).}
 \resizebox{0.45\textwidth}{!}{
\begin{tabular}{l|llllll}
\hline
            & Barlow Twins & BYOL & SimSiam & SimCLR & DINO & SWaV \\ \hline
Batch Size  & 2048         & 4096 & 256     & 4096   & 1024 & 256  \\
Epochs      & 300          & 200  & 100     & 200    & 800  & 200  \\ \hline
Linear-Eval\%   & 71.8         & 71.8 & 68.3    & 66.9   & 75.3 & 70.5 \\\hline 
Epoch       & 90           & 90   & 90      & 90     & 100  & 100  \\
Batch Size  & 256          & 512  & 512     & 512    & 256  & 256 \\
\hline
\end{tabular}
}
    \label{tab:config}
\end{table}
\textbf{Experimental details} for evaluating the robustness of self-supervised representations are as follows. We extracted the encoder from a ImageNet pre-trained self-supervised representation learning model and added a classification layer at the end of the network. This allows the model to be fine-tuned on ImageNet 2012 dataset for a classification task. Evaluations is performed on ImageNet-C dataset~\cite{imagenetc}. For this work, we have considered six of the state-of-the-art SSL algorithms, and the configurations are shown in Table~\ref{tab:config}.
We first initialized the classifier layer randomly and froze all the parameters of the pre-trained encoder. Next, we trained the classifier using the labeled training set. 
The models used were trained by mmsetup\cite{mmselfsup2021} except for DINO, which came from its original repository \cite{dino}.
ResNet-50 was chosen across all different methods to keep the analysis uniform, and all experiments subsequently were conducted using this architecture. The SSL models were tested on ImageNet-C~\cite{imagenetc}, and $mCE$~\cite{imagenetc} is used as a performance measure. The results are shown in Table \ref{tab:overview} and \ref{tab:overview_large}. 
%


\section{Can SSL methods endure shifts in data distribution and image corruptions?}
\label{sec:RQ}
The raised research questions are discussed in this section.

\textbf{Q1: How do self-supervised representation learning (SSL) paradigms (contrastive learning, knowledge distillation, mutual information maximization, clustering) perform in terms of robustness when exposed to distribution shifts and image corruptions?} \label{Q1}

\begin{figure}
     \centering
     \includegraphics[width=0.4\textwidth]{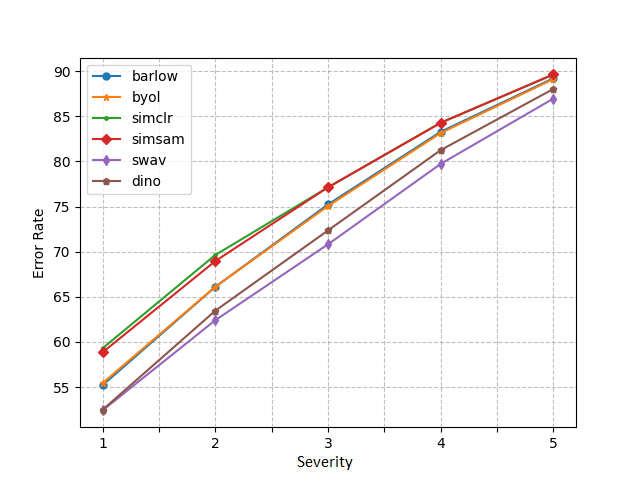}
     \caption{Error rates vs. severity levels across ImageNet-C \cite{imagenetc} corruptions. }
     \label{fig:avg}
 \end{figure}

The average error rates against all corruptions (per severity level) of all the SSL methods are depicted in Figure~\ref{fig:avg}. The general trend is that SimCLR and SimSam have higher error rates as compared to other methods. While  learning has reported good performance previously on ImageNet-C \cite{khosla2021supervised}, we noticed that SimCLR is not comparably robust against these corruptions. 
A pattern observed (Figure \ref{fig:avg}) is that, in general, Knowledge distillation methods seem to outperform contrastive learning. Clustering outperforms other methods indicating robust representations. From Figure~\ref{fig:avg}, one important observation is that for corruptions with lower severity levels, the six SSL methods form three sets where SwaV and DINO perform best, followed by BYOL and Barlow twins; finally, SimCLR and SimSiam have relatively lower performance. However, at the highest severity level, all the methods have similar and high error rates. This is likely because most images in this group are heavily distorted and challenging even for the human visual system to comprehend.
From Table~\ref{tab:overview}, we observe that SwaV outperforms all the competing methods in terms of corruption error and mean corruption error; however, DINO has a better robustness performance. 
\begin{table*}[ht]
    \centering
    \caption{ Results for each method calculated over the corruption metric \cite{imagenetc}.}
    \begin{tabular}{l|rrrrrr}
        \hline
        {} &     Barlow Twin &       BYOL &     SimCLR &     SimSam &       SWaV &       DINO \\
        \hline
            clean error &   28.2 & 28.2 &   33.1 &   31.7 & 29.5 & \textbf{24.7} \\
            average error    &   73.8 & 73.8 &   75.99 &   75.8 & \textbf{70.5} & 71.5 \\
            average relative error        &   74.7 & 74.6 &   76.0 &   76.0 & \textbf{70.7} & 72.9 \\ \hline
    \end{tabular}
    \label{tab:overview}
\end{table*}


\textbf{Q2: To what extent can self-supervised representation learning methods maintain their robustness in the presence of distribution shifts, and what are the factors that limit their ability to do so?}\label{Q2}

\begin{figure*}[ht]
    \centering
    \includegraphics[width=0.95\textwidth]{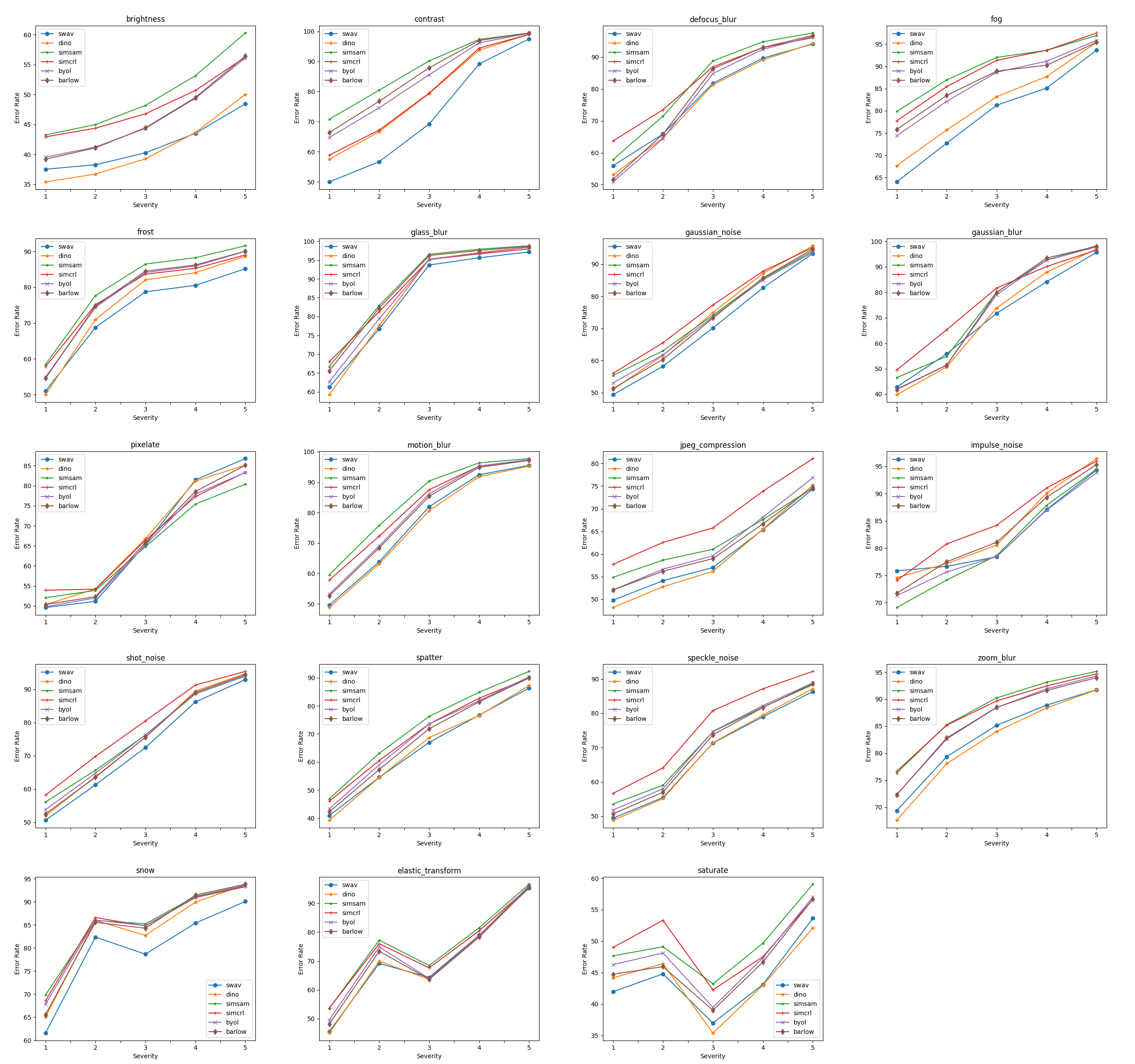}
    \caption{Model performance against specific corruptions by severity. For corruptions, namely, snow, saturate, and elastic (last row), SSL models perform poorly at severity level 2 than at severity level 3.}
    \label{fig:3-elst-sat-sno}
\end{figure*}

Table~\ref{tab:overview_large} presents a detailed analysis using mean corruption error $mCE$ for each corruption. Here, we report the average $mCE$ for each corruption in the ImageNet-C dataset. One of the findings is that glass blur significantly impacts the robustness of these models, specifically at higher severity levels. Most of these models have demonstrated good robustness to brightness-based corruptions. As corroborated by Figure~\ref{fig:3-elst-sat-sno} for most corruptions, the model robustness suffers with the increase in severity levels.

\begin{table*}[h!]
    \centering
        \caption{$mCE$ for each corruption type against the baseline. The error rates in each column of corruption types are average values of all severity levels.}
    \resizebox{1\textwidth}{!}{
   \begin{tabular}{lrrrrrrrrrrrrrrrrrrrr}
                                         & \multicolumn{1}{l}{}              & \multicolumn{4}{l}{\textbf{Weather}}                                        & \multicolumn{3}{l}{\textbf{Noise}}                             & \multicolumn{4}{l}{\textbf{Extra}}                                                             & \multicolumn{4}{l}{\textbf{Digital}}                                                          & \multicolumn{4}{l}{\textbf{Blur}}                                   \\ \hline
\multicolumn{1}{l|}{}                    & \multicolumn{1}{r|}{\textbf{mCE}} & \textbf{Snow} & \textbf{Frost} & \textbf{Fog} & \multicolumn{1}{r|}{\textbf{Bright}} & \textbf{Gauss.} & \textbf{Shot} & \multicolumn{1}{r|}{\textbf{Impulse}} & \textbf{Speckle} & \textbf{Gauss.} & \textbf{Spatter} & \multicolumn{1}{r|}{\textbf{Saturate}} & \textbf{Pixelate} & \textbf{Contrast} & \textbf{Elastic} & \multicolumn{1}{r|}{\textbf{JPEG}} & \textbf{Zoom} & \textbf{Defocus} & \textbf{Motion} & \textbf{Glass} \\ \hline
\multicolumn{1}{l|}{\textbf{barlow}}     & \multicolumn{1}{r|}{73.8}         & 84            & 78             & 87           & \multicolumn{1}{r|}{46}     & 73              & 75            & \multicolumn{1}{r|}{83}      & 70               & 73              & 69               & \multicolumn{1}{r|}{47}                & 66                & 85                & 72               & \multicolumn{1}{r|}{62}            & 86            & 79               & 80              & 88             \\
\multicolumn{1}{l|}{\textbf{byol}}       & \multicolumn{1}{r|}{73.8}         & 85            & 78             & 86           & \multicolumn{1}{r|}{46}     & 73              & 76            & \multicolumn{1}{r|}{81}      & 71               & 73              & 70               & \multicolumn{1}{r|}{48}                & 66                & 84                & 73               & \multicolumn{1}{r|}{63}            & 86            & 78               & 80              & 86             \\
\multicolumn{1}{l|}{\textbf{simclr}}     & \multicolumn{1}{r|}{76.0}         & 85            & 78             & 89           & \multicolumn{1}{r|}{48}     & 76              & 79            & \multicolumn{1}{r|}{85}      & 76               & 77              & 70               & \multicolumn{1}{r|}{50}                & 67                & 80                & 75               & \multicolumn{1}{r|}{68}            & 88            & 83               & 82              & 88             \\
\multicolumn{1}{l|}{\textbf{simsam}}     & \multicolumn{1}{r|}{75.8}         & 85            & 80             & 90           & \multicolumn{1}{r|}{50}     & 74              & 76            & \multicolumn{1}{r|}{81}      & 71               & 74              & 73               & \multicolumn{1}{r|}{50}                & 65                & 88                & 75               & \multicolumn{1}{r|}{63}            & 88            & 82               & 84              & 89             \\
\multicolumn{1}{l|}{\textbf{swav}}       & \multicolumn{1}{r|}{70.5}         & 80            & 73             & 79           & \multicolumn{1}{r|}{41}     & 71              & 73            & \multicolumn{1}{r|}{82}      & 68               & 70              & 65               & \multicolumn{1}{r|}{44}                & 67                & 73                & 71               & \multicolumn{1}{r|}{60}            & 83            & 77               & 77              & 85             \\
\multicolumn{1}{l|}{\textbf{dino}}       & \multicolumn{1}{r|}{72.9}         & 83            & 75             & 82           & \multicolumn{1}{r|}{41}     & 74              & 75            & \multicolumn{1}{r|}{84}      & 68               & 70              & 65               & \multicolumn{1}{r|}{44}                & 68                & 79                & 70               & \multicolumn{1}{r|}{60}            & 82            & 76               & 76              & 85             \\
\multicolumn{1}{l|}{\textbf{supervised~\cite{imagenetc}}} & \multicolumn{1}{r|}{76.7}         & 78            & 75             & 66           & \multicolumn{1}{r|}{57}     & 80              & 82            & \multicolumn{1}{r|}{83}      & 76               & 74              & 76               & \multicolumn{1}{r|}{58}                & 77                & 71                & 85               & \multicolumn{1}{r|}{77}            & 80            & 75               & 78              & 89          \\ \hline
\end{tabular}
    }
    \label{tab:overview_large}
\end{table*}

{
\textbf{Q3: What is the relationship between the robustness performance of different SSL paradigms and common categories of corruptions? }\label{Q3}

As the severity levels of corruptions increase, all self-supervised representation learning (SSL) methods demonstrate a decline in their robustness, as shown in Figure~\ref{fig:group}. While the \textit{noise} and \textit{blur} groups have a similar trend, whereas  \textit{digital} group shows comparatively strong resilience for intermediate severity level. SSL methods are least robust against \textit{weather} group.

\begin{figure*}
    \centering
    \includegraphics[width=0.9\textwidth]{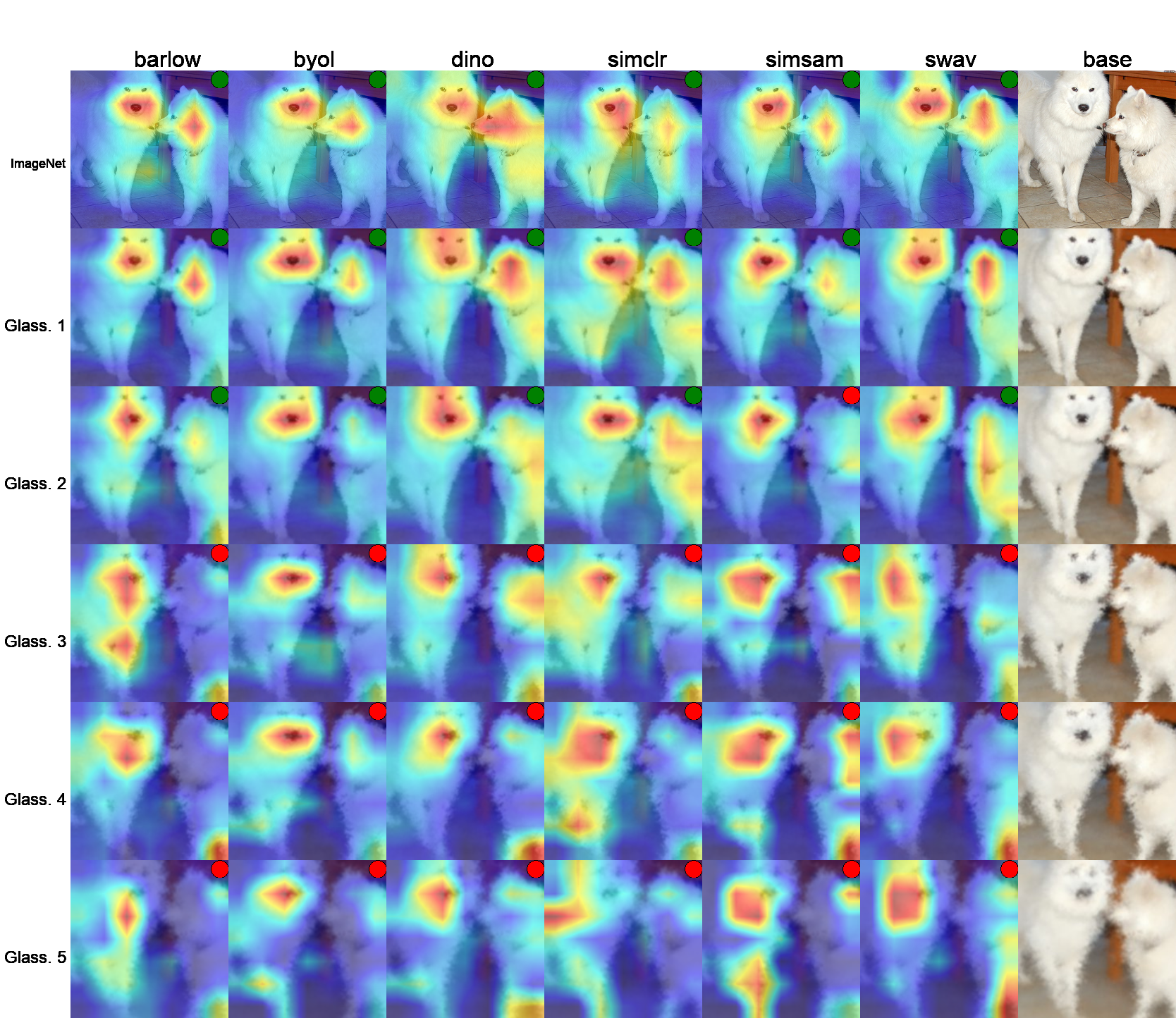}
    \caption{Glass blur on dogs; markers in the images show correct (green) and incorrect (red) classifications. In ImageNet, with many dog breed classes, misclassification doesn't necessarily indicate a bad model if the representation is adequate. In the twin dog example, with low blur severity, both dogs have good activations for all models, suggesting good representations. However, at high blur severity, the model struggles to classify, resulting in distorted activations and difficulty in distinguishing between the dogs, leading to poor results.}
    \label{fig:grad_dog_grid}
    \centering
        \includegraphics[width=.38\linewidth]{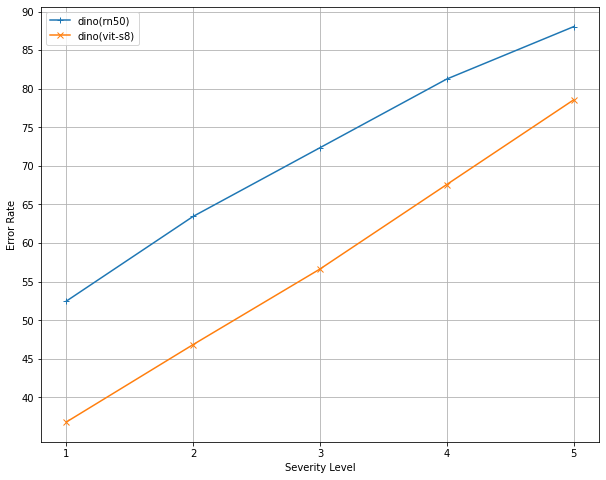}
        \includegraphics[width=.45\linewidth]{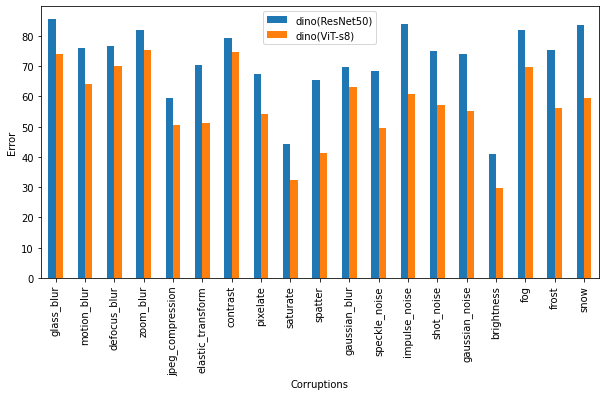}
    \caption{Comparison between different backbones, ResNet50 and ViT-s8 for DINO SSL method over ImageNet-C \cite{imagenetc} corruptions. Severity levels (left), corruptions (right).}
    \label{fig:dino_compare_ic}
\end{figure*}


\begin{figure*}[htbp]
    \centering
        \includegraphics[width=.22\linewidth]{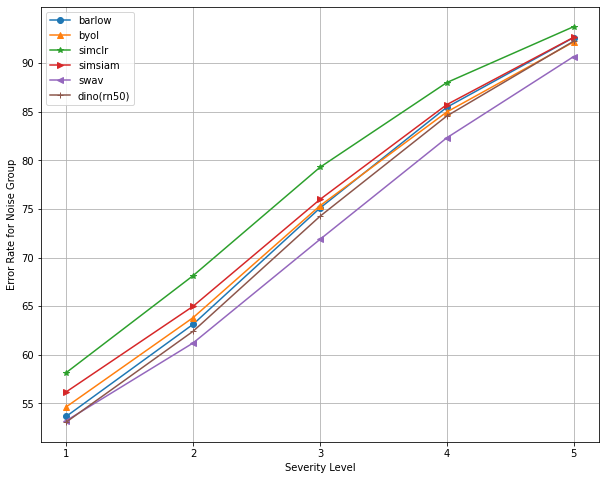}   
        \includegraphics[width=.22\linewidth]{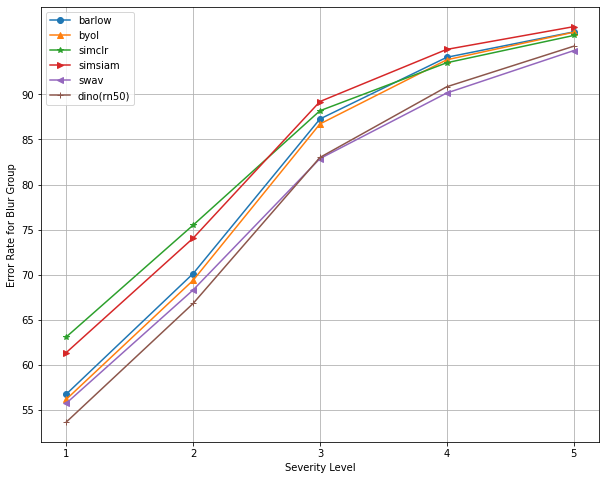}
        \includegraphics[width=.22\linewidth]{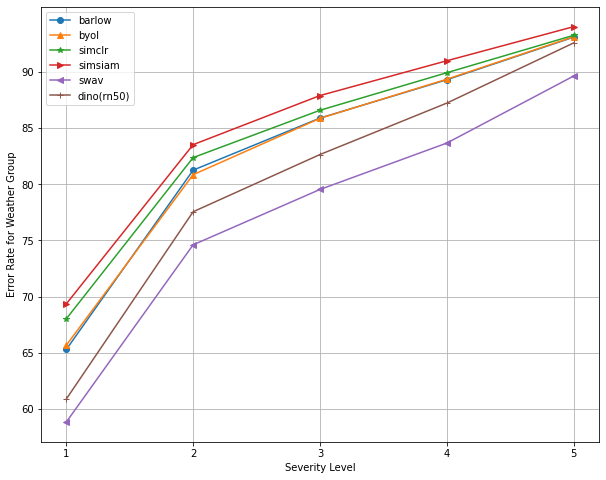}
        \includegraphics[width=.22\linewidth]{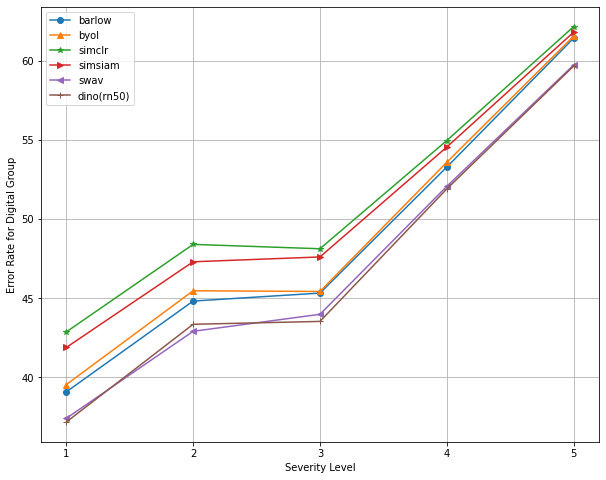}
    \caption{Group-wise comparison. (a) Noise (b) Blur (c) Weather (d) Digital (left to right).}
    \label{fig:group}
\end{figure*}

{
\textbf{Q4: Do self-supervised representation learning methods deviate from the observed trend of error increase for certain corruptions, and what factors contribute to their robustness in the face of these corruptions?} \label{Q4}

We observed (Figure~\ref{fig:3-elst-sat-sno}) that for three corruptions, namely, \textit{snow}, \textit{saturate}, and \textit{elastic transform} (last row), there is a deviation from the expected trend; the expected trend is that the error increases with an increase in severity level. However, SSL models are performing low at severity level 2 than at severity level 3. 
Given the intriguing deviations displayed by (snow, elastic, saturate) from their anticipated behavior, we delved deeper into our inquiry, employing a renowned perceptual measure known as
Structural Similarity Index measure (SSIM)~\cite{struct} to investigate further, as one of the metrics popularly used by image quality researchers for reference image-based quality assessment.

We computed the SSIM between the original image (from ImageNet) and the corresponding corrupted image (from ImageNet-C) for all test images and averaged at each severity level (Table \ref{tab:ssim}); this gives an estimate of the visual quality. 

\textit{Snow} corruption occludes the object by adding whitish pixels as snowflakes with motion blur. It has more visually challenging images at severity level 2 than other levels; therefore, SSIM at level 2 is lower than the SSIM at other severity levels. 
Similarly, for \textit{elastic transform}, the SSIM at level 2 is lower than the SSIM at other severity levels. At low severity (levels 1 and 2), the affine transform is more noticeable in some cases, causing artifacts, which can also be seen in figure \ref{fig:grad_rep_grid} on its elastic transform. 
The \textit{saturation} corruptions have very low saturation at low severity levels, causing it to be a grayscale image. This might lead to some classes not being accurately predicted, where color information is crucial. In nutshell, only for snow, elastic and saturate, increased severity level (2 to 4) by increasing respective artifacts, does not reflect increased noise in image examples which mitigates above stated behaviour from all SSL methods.

 \begin{table}[h]
\centering
    \caption{SSIM metric for snow, elastic and saturate-based corruptions.}
    
\begin{tabular}{l|lll}
\hline
Severity & Snow            & Elastic    & Saturate            \\ \hline
1    & 0.218 & 0.276  & 0.288 \\
2    & 0.179  & 0.237 & 0.283  \\
3    & 0.194  & 0.315  & 0.273  \\
4    & 0.186  & 0.312  & 0.234  \\
5    & 0.189 & 0.305  & 0.210 \\ \hline
\end{tabular}
    \label{tab:ssim}
\end{table}


\textbf{Q5: To what extent does the presence of corruptions shift the focus of classifiers from overall representation to specific features?}\label{Q5}

To gain more insight into how different self-supervised representation learning methods for classification task pick a label, we have used gradcams~\cite{gradcam} to compare the different methods qualitatively. Gradcams are used to explain the model's decision as they provide heatmaps on where in the image the model is focusing. 
In Figure~\ref{fig:grad_rep_grid}, we show the grad cams of an image for all SSL methods under different corruptions of varying severity levels. 

The difference among Gradcams gives an understanding of how the model behavior changes in the presence of a particular corruption. From Table~\ref{tab:overview_large}, we noticed that \textit{glass} blur corruption had caused the highest misclassification for all competing SSL methods; to understand how different methods respond to different severity of \textit{glass} blur, we provide the corresponding gradcams in Figure~\ref{fig:grad_dog_grid}.
\begin{figure*}
        \centering
    \includegraphics[width=1\textwidth]{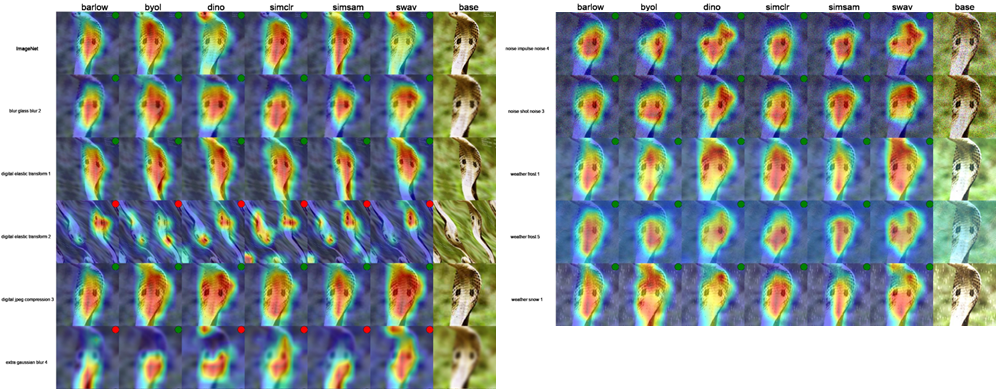}
    \caption{Random Corruptions on a Cobra; the markers in the images show correct and incorrect classifications. Cobra, a reptile with multiple classes in ImageNet, may confuse classifiers. However, cobras are generally distinguishable from other reptiles due to their distinctive neckband. At severity level 2 of elastic transform, there are artifacts causing a distorted doublet, making it a challenging case. Overall, models perform well with good representation. The classifier in these examples shows a bias toward the neckband, while the original un-cropped image emphasizes the edges for classification by DINO and SwaV. However, due to corruptions, the focus shifts more towards the neckband.}
    \label{fig:grad_rep_grid}
\end{figure*}

\textbf{Q6: Do different backbones, such as Convolutional Neural Networks (CNNs) and Transformers, influence the behavior and robustness?}\label{Q6}

There has been analysis~\cite{heo2023exploring} on adversarial robustness for transformer and CNN architectures but to specifically analyze the robustness against corruptions and distribution shifts, we chose the most robust SSL method from the previous analysis (i.e., DINO), and compared the backbone ViT-s8~\cite{caron2021emerging} transformer with standard CNN ResNet-50.  Undoubtedly, transformer architecture outperformed CNN backbone across the severity levels and also for each image corruption. A detailed trend is shown in Figure~\ref{fig:dino_compare_ic}. 
\section{Conclusion}
The primary objective of this investigation was to conduct an in-depth analysis of diverse paradigms employed in current self-supervised representation learning paradigms, focusing on their robustness characteristics when subjected to varying corruptions present in the ImageNet-C database. The aim was to gain a comprehensive understanding of how these self-supervised representation learning paradigms perform and behave in the face of diverse corruptions, thereby contributing to the advancement of robust representation learning in the computer vision domain. Through empirical analysis, we have presented various analytical trends and demonstrated that self-supervised representation learning methods exhibit decreased robustness as distributional shifts intensify. Notably, our findings indicate that the DINO method employing the distillation approach and the SwAV method utilizing clustering exhibit relatively higher levels of robustness compared to the other methods investigated in this study. While DINO is associated with knowledge distillation, SwAV employs a contrastive assignment quantization approach, indicating their dissimilarity in methodology. These results suggest that multiple SSL methods originating from diverse SSL paradigms display enhanced robustness when evaluated on ImageNet-C. However, it is essential to view these empirical findings as a starting point for further exploration rather than definitive conclusions.
The comparative study conducted in this research serves to enhance the comprehension of the computer vision community regarding the strengths and limitations of various self-supervised representation learning approaches. Furthermore, it facilitates researchers in developing robust representations in future endeavors. A significant finding from our analysis is that the SwaV method, which employs a clustering approach, exhibits higher robustness compared to popular methods such as SimCLR and Barlow Twins. This result offers valuable insights for future research directions aimed at further improving self-supervised representation learning methodologies. Considering the findings of this study, it becomes imperative to address the challenges associated with the performance degradation of self-supervised representation learning methods under distribution shifts and image corruptions. By prioritizing safety and robustness, researchers can contribute to the development of more reliable and trustworthy self-supervised representation learning techniques that can effectively handle real-world scenarios and enhance the practical utility of these methods.
In this work, we dedicate to the methodical revelation of empirical evidence, rather than hypothesizing. Our endeavor remains steadfast in illuminating numerous enigmas through a rigorous examination. We firmly hold the conviction that this pioneering work shall pave the way for future inquiries, enabling the formulation and evaluation of cogent hypotheses. 

\bibliographystyle{ieee_fullname}
\bibliography{egbib}
\end{document}